\documentclass[conference, a4paper]{IEEEtran}

\ifCLASSOPTIONcompsoc
    \usepackage[caption=false,font=normalsize,labelfont=sf,textfont=sf]{subfig}
\else
  \usepackage[caption=false,font=footnotesize]{subfig}
\fi
\usepackage{cite}
\usepackage[pdftex]{graphicx}

\begin{document}

\title{An Embedded Decision Support System for Runway Safety and Excursion Avoidance}

\author{

\IEEEauthorblockN{
Georgios Alogdianakis\IEEEauthorrefmark{1}, 
Ioannis Katsidimas\IEEEauthorrefmark{1}, 
Athanasios Kotzakolios\IEEEauthorrefmark{1},\\ 
Anastasios Plioutsias\IEEEauthorrefmark{2}, 
Vassilis Kostopoulos\IEEEauthorrefmark{1}
}

\IEEEauthorblockA{\IEEEauthorrefmark{1} Applied Mechanics Laboratory, Department of Mechanical Engineering \& Aeronautics, University of Patras}
\IEEEauthorblockA{\IEEEauthorrefmark{2}EEC School of Future Transport Engineering (FTE),        Coventry University}

\IEEEauthorblockA{
up1055737@ac.upatras.gr, 
ikatsidima@ceid.upatras.gr, 
kotzakol@upatras.gr, \\
tassos.plioutsias@coventry.ac.uk, 
kostopoulos@upatras.gr
}

}

\maketitle

\begin{abstract}
Runway Safety Assistant Foreseeing Excursions (RUN.S.A.F.E.) is a complete embedded system solution that predicts a potential runway overrun during the takeoff and landing of a civil aviation aircraft. The system executes both static and dynamic calculations, the former being completely dependent, while the latter completely independent to the user’s inputs. The solution is adapted to a Boeing 737-800 aircraft, with CFM56-7B engines. However, the calculations also apply for similar aicrafts, equipped with a tricycle landing gear and turbofan engines. The system is aligned with current regulations and certification specifications, where applicable.
\end{abstract}

\begin{IEEEkeywords}
Aviation safety, Avionics, Runway safety, Embedded system, Real time system, Human Computer Interaction
\end{IEEEkeywords}

% \fi
\IEEEpeerreviewmaketitle

\section{Introduction}
Take off and landing procedure are two of the most critical parts of a commercial aircraft's flight. Runway overrun incidents, i.e. when an aircraft exceeds the runway's length, despite the low mortality rate, can lead to passenger confusion, runway and aircraft damage, etc. This type of incident is the most common runway related accident in recent years \cite{iata_runway_accident_analysis}. 

Human factor leads the causality list of aviation accidents \cite{kharoufah_review_2018}, thus a system that advises and warns the operators in cases of a potential runway overrun is a necessity, to decrease the risk and occurrence of such incidents. The critical speeds of the two procedures (\textit{V1}, \textit{VR}, \textit{V2} for takeoff and \textit{Vapp}, \textit{Vref} for landing) are calculated through manual or specialised software solutions (e.g. Flysmart+ \cite{flysmart}), and aim to prevent a possible overrun, while ensuring convenient acceleration/deceleration rates. 
%However, key assumptions, like optimal engine performance and runway friction, may not always be valid, contributing to the causing factors of these incidents. 
%Pilots mainly rely on experience and visual clues to estimate the remaining runway, as instruments only provide speed indications with accuracy. Recognizing these limitations, there's a call for an auxiliary runway excursion safety system in aircraft cockpits. The Runway Safety Assistant Foreseeing Excursion (RUNSAFE) system performs static calculations based on user inputs and dynamic calculations during each procedure's development. Aerodynamic and performance data for these calculations came from the X-Plane 11 flight simulator.
Pilots mainly rely on experience and visual clues to estimate the remaining runway, as instruments can only provide speed indications with accuracy. 

%A prototype embedded system was developed on a RaspberryPi, in order to examine the feasibility, efficiency and integration of the system within resource constrained hardware. A seven inch touch screen and speakers were attached to it for input and output interface purposes. 

\section{Motivation and State of the Art}
The development of a Take Off Performance Monitor System (TOPMS) is initially established by Aerospace Standard 8044 in 1987 \cite{as8044}.  While NASA developes prototypes \cite{middleton1994flight}, currently there are no complete systems installed in commercial aircrafts, only detectors of major errors between the calculated and actual acceleration during take off \cite{airbus2020safety}. In EASA's (European Union Aviation Safety Agency) research agenda for 2022-2024, Reference LOC-03, named "Landing and take-off monitor" requests to "assess means to assist the flight crew in preventing runway overrun and managing aircraft total energy, as well as monitoring the actual acceleration of the aircraft during the take-off run to detect mismatch between \textit{V1} and the actual remaining runway distance" \cite{easa_research_agenda_2022}. 

EASA's Certification Specifications for large aircrafts (CS-25) require a type of Runway Overrun Awareness and Alerting System (ROAAS), to be installed in all new aircrafts in order to comply with the rules. This system "shall reduce the risk of a longitudinal runway excursion during landing by providing alert, in flight and on ground, to the flight crew" \cite{cs_25_am_27}. Two of the greatest commercial aircraft manufacturers, Airbus and Boeing, have developed their own solutions named Runway Overrun Prevention System (ROPS) and Runway Awareness and Advisory System (RAAS) respectively. Other prototypes have been developed for the Gulfstream G650 \cite{g650_roaas}. 

Most of aviation relevant research works regarding CPS and embedded systems go beyond overrun, offering services and solutions based on SHM and RUL applications which exclude human factor. As such, Athanasakis et.al. present an embedded intelligence solution for the RUL of a turbofun \cite{Athanasakis}, while other studies explore promising impact detection \cite{smartobject}, damage detection and severity estimation in SHM use cases \cite{damage_detection} and predictive maintenance operations \cite{predictive_maintenance}. 

\section{Models and calculations}
The system calculates the critical distances \textit{Accelerate Stop Distance Required}, \textit{ASDR}, for takeoff and \textit{Landing Distance Required}, \textit{LDR}, for landing, in correspondence with the critical speeds \textit{V1} and \textit{Vref}, respectively. Two types of calculations are performed: a static which depends on the user's inputs, and a real time, based on the aircraft's acceleration and speeds.

\subsection{Take Off Procedure}
\begin{figure}[!t]
\centering
\includegraphics[width=0.8\linewidth]{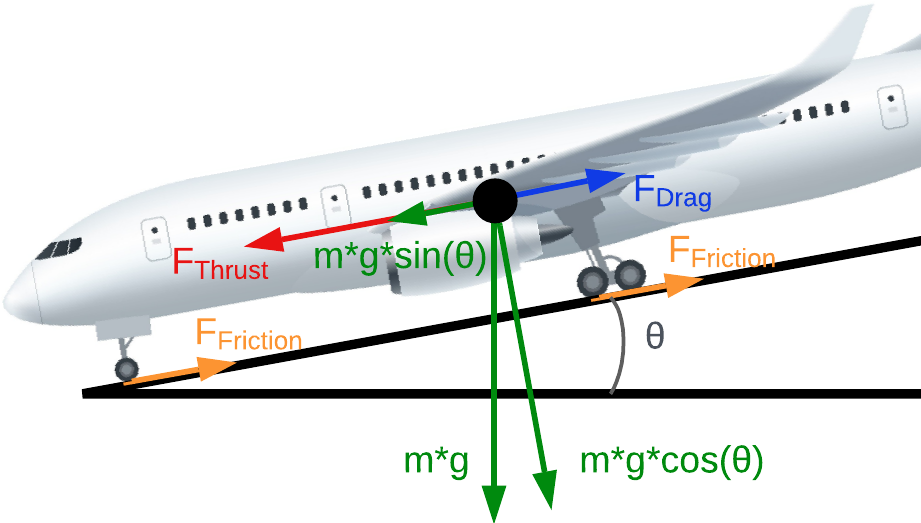}
  \caption{Forces acting on the aircraft while on the runway. }
  \label{fig:forces}
\end{figure}

\paragraph{Static Calculations}
During takeoff and before the aircraft is lifted from the ground, there is a possibility of events (e.g. engine failure), that require the initiation of a Rejected Take Off procedure. 
%In this case, a series of actions has to be made from the pilots to bring the aircraft to a stop within the runway's limits. 
This action is triggered only prior to the \textit{V1} speed, above which  the remaining runway distance is not enough to bring the aircraft to a stop. To statically calculate the \textit{ASDR}, we model \textit{Thrust}, \textit{Drag}, \textit{Friction} and \textit{Gravitational Force} (due to the runway's slope), as seen in Figure \ref{fig:forces}.

% To statically calculate the ASDR, all the forces acting on the longitudinal axes are modeled. Those include Thrust, Drag, Friction and Gravitational Force (due to the runway's slope), as seen in Figure \ref{fig:forces}. 

%Each of the forces is modeled to an adequate level of precision that allows reasonable deviations at the final distance result. 
Minor differences exist in the accelerating and decelerating part of the calculations, which aim to maintain low complexity levels. By knowing the aircraft's mass, we  find the acceleration at each moment based on Newton's second law $\sum\vec{F} = m\vec{a}$. 

\begin{figure}[!t]
\centering
\includegraphics[width=\linewidth]{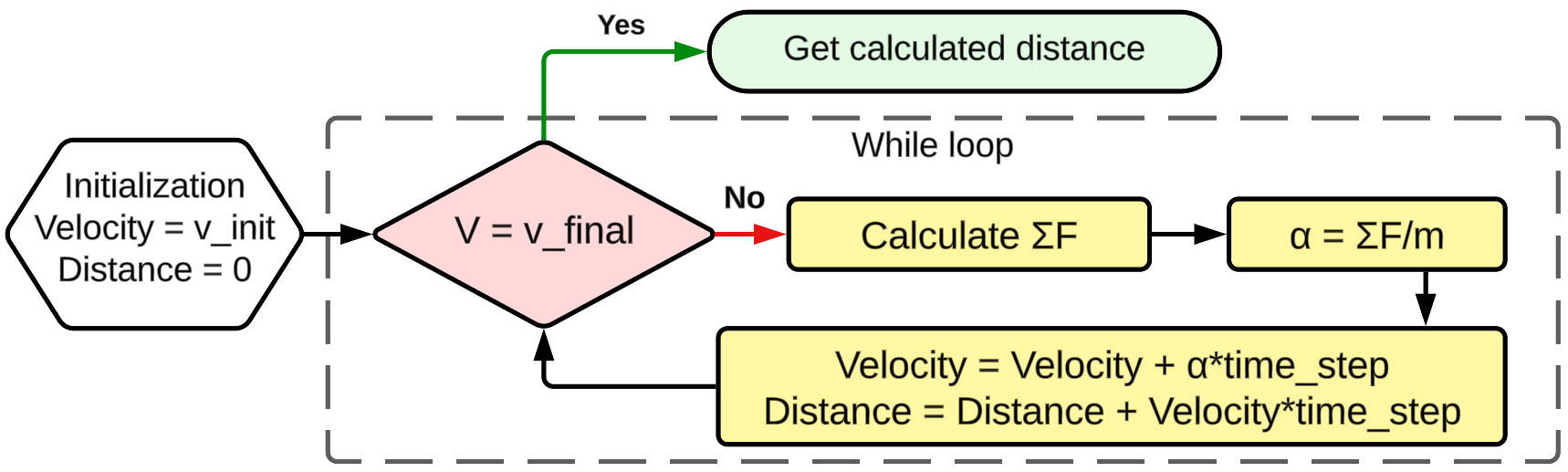}
  \caption{While loop for calculating distances.}
  \label{fig:while_loop}
\end{figure}

Using a double integral on the acceleration, 
the distance covered by the plane to reach
the desired speed, is calculated. The code that implements these 
integrals is based on
the Riemann sum, in which the definite integral is computed as a sum of 
area of parallelograms.
The independent variable of the function and integrals is time, while 
the division also defines
the precision of the integral. The flowchart in Figure \ref{fig:while_loop} shows the structure of the algorithm.

The selected time step determines the accuracy rate of the final result and the number of the required operations. For this particular 
implementation, the golden medium between the two, is set at
0.05 seconds(after numerous experiments), and leads to almost instant calculations, without sacrificing the accuracy ($<$1m in the majority of the cases).

\begin{figure}[!t]
\centering
\includegraphics[width=0.8\linewidth]{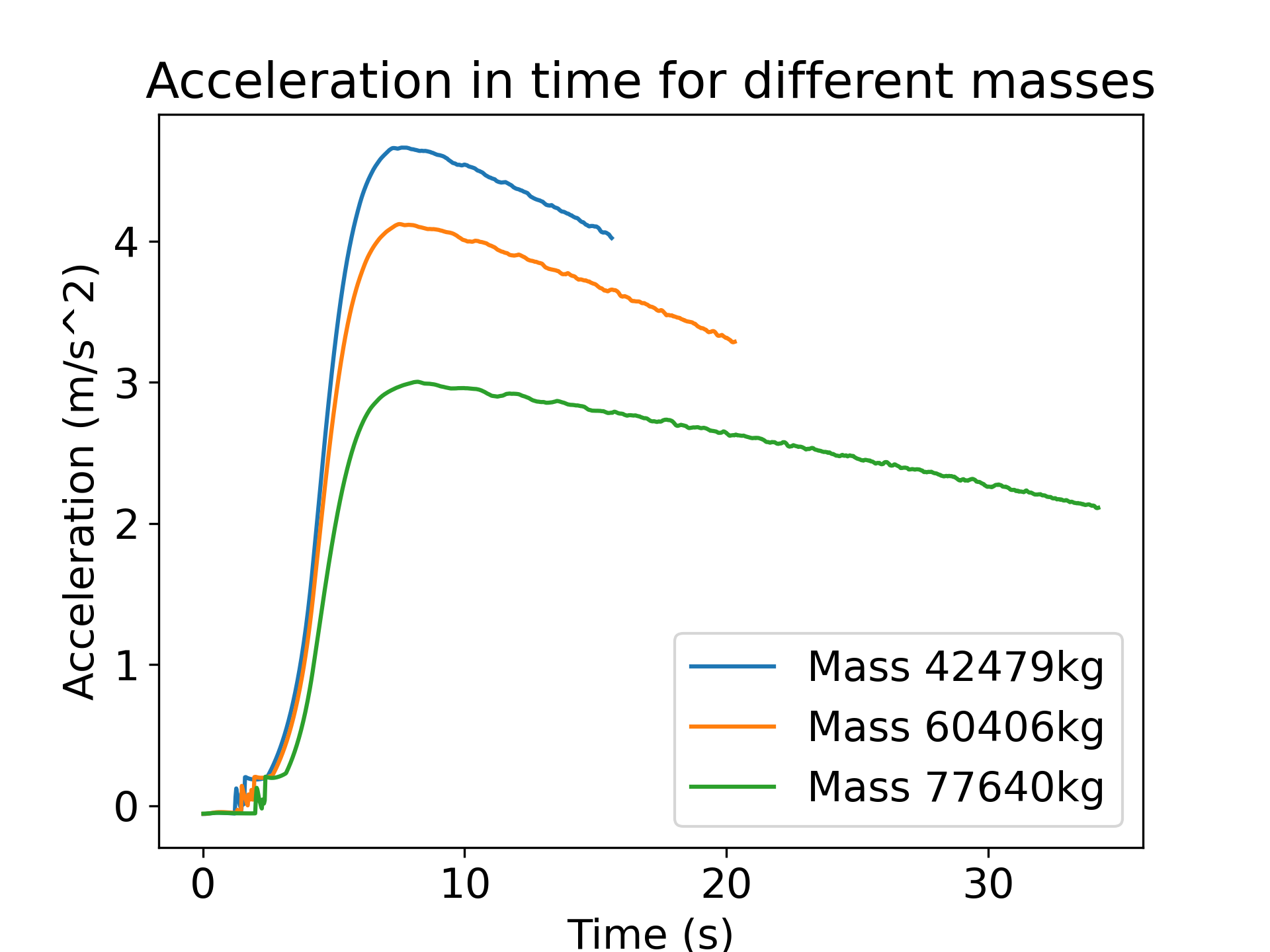}
  \caption{Acceleration during a take off of the Boeing 737-800 in the X-Plane simulator.}
  \label{fig:737_sim_acc}
\end{figure}

\paragraph{Dynamic Calculations}
%After the completion of the static calculations, the system is ready for the dynamic ones. 
The procedure is initiated when the thrust lever is moved from the idle position. In Figure \ref{fig:737_sim_acc} the acceleration in time is presented during three different configurations, resulting in a linear negative correlation between the acceleration and time, that is also observed in practice \cite{middleton1994flight}.

Data received between 8th and 10th second after the initiation of a takeoff, are used in a Linear Regression to determine the coefficients $\alpha, \beta$ that describe the acceleration as a linear equation, $acceleration(t) =  \alpha*t + \beta$. These coefficients are used as seeds for the next algorithm, namely Recursive Least Squares (RLS) algorithm. The latter estimates the coefficients of acceleration aiming to minimize a weighted least square cost function in relevance to the input signal. The forgetting factor is set to 1, hence the system behaves as a linear regression in which each past input contributes equally to the least squares cost function.  

By integrating the acceleration function in time, the velocity is calculated, as $velocity(t) = \frac{\alpha}{2}t^2 + \beta t + \gamma$, in which $\alpha, \beta$ are the coefficients of velocity function and $\gamma$ is the speed directly after the spoolup completion. The time at which the desired speed is attained, can be found from this second degree function. By integrating once more, to find the distance function in time and replacing with the previous result, the distance covered till the attainment of the velocity can be found. Similar logic applies to the dynamic calculations of the decelerating part. The only difference is that the RLS algorithm approximates the coefficients $\alpha, \beta, \gamma $ of the velocity function in time. 
%When applying full pressure to the brakes, the anti skid system is activated, leading to fluctuations of the deceleration rate during braking. These fluctuations would set the convergence of the RLS algorithm impossible, hence the alteration. 

\begin{figure}[!t]
\centering
\includegraphics[width=\linewidth]{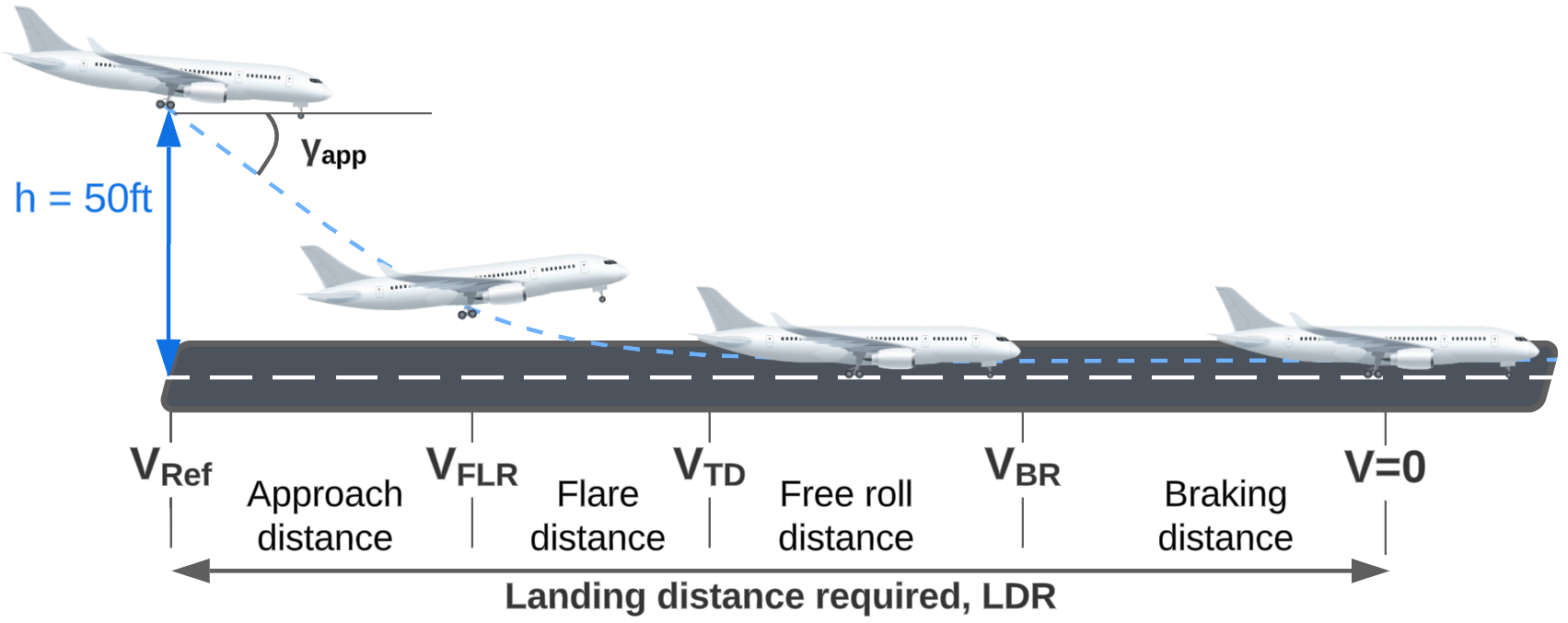}
  \caption{Phases of landing. }
  \label{fig:landing_phases}
\end{figure}

\subsection{Landing Procedure}
The landing procedure can be divided into 4 distinct phases: $Approach$, $Flare$, $Free\ Roll$ and $Braking$, as seen in Figure \ref{fig:landing_phases}. The Approach stage starts 50ft above the runway where the aircraft is considered to be exactly above the threshold of the runway, the $Flare$ stage starts at about 20 feet, the $Free\ roll$ when the aircraft touches the ground and the $Braking$ when the auto brake system is activated or the pilots press the pedals. 

\paragraph{Static calculations} When approaching the runway the aircraft has a steady glide path of 3 degrees and a steady precalculated \textit{Vref} speed. This is enough to calculate the distance traveled between 50 and 20 feet. The Flare stage is a peculiar and hard to model maneuver, aiming to reduce the vertical speed before coming in contact with the runway. The mean speed during this stage is 98\% of the \textit{Vref} speed. The Free roll stage lasts 3.5 seconds on average, at a speed reduced by 7.5\% in comparison to \textit{Vref}.  Finally, Braking lasts till the aircraft's velocity is 0, at a steady deceleration rate by the Autobrake, which directly gives the corresponding distance. 

\paragraph{Dynamic calculations} The dynamic calculations also include a pre-approach state, starting at 300ft above the runway level, to provide timely notifications. The first three stages of dynamic calculations, \textit{Pre-Approach}, \textit{Approach} and \textit{Flare}, use the moving average of the horizontal and vertical speed. 
Based on the vertical speed and the height difference till the lower limit of the stage, the time needed till completion is calculated. By multiplying with the horizontal speed, the horizontal distance till completion is also calculated. The three stages have different sensitivities to changes, with the last being the most sensitive.
The \textit{Free Roll} stage utilizes the moving average of the horizontal speed solely and it is multiplied with time left till the completion of the stage (considered to be 3.5 seconds), to find the distance left to be covered at this stage. Lastly, the braking stage uses the RLS algorithm, taking advantage of the steady deceleration rate. 

\section{System}
\subsection{Hardware}
\begin{figure}[!t]
\centering
\subfloat[]{\includegraphics[width=0.49\columnwidth]{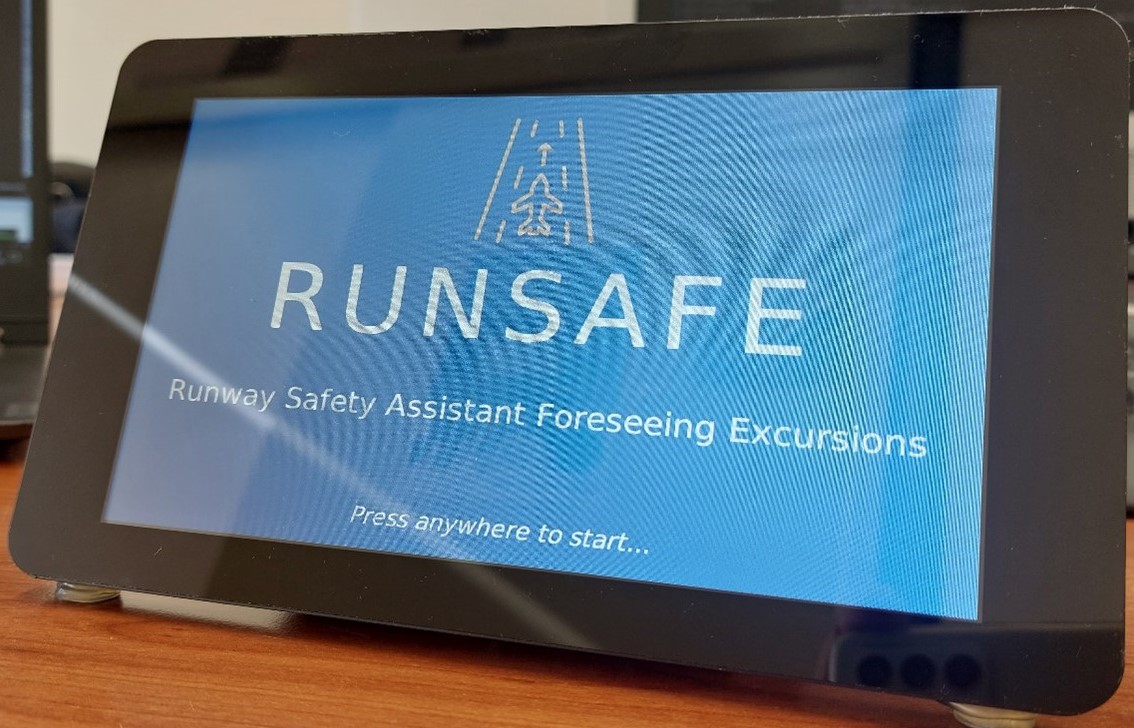}%
\label{fig:welcome_page_photo}}
\hfil
\subfloat[]{\includegraphics[width=0.45\columnwidth]{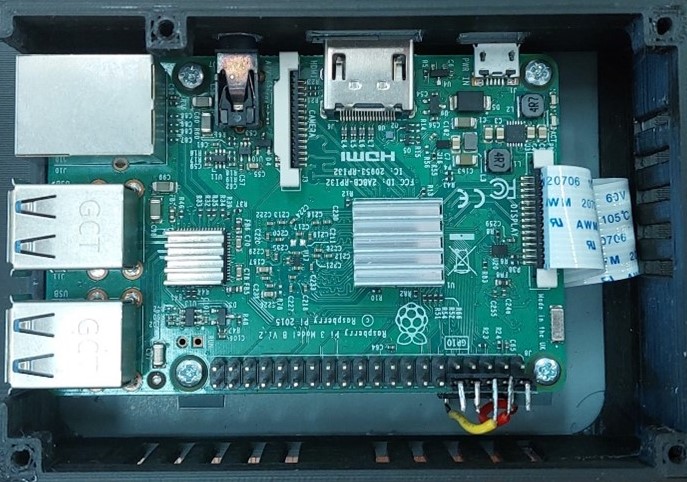}%
\label{fig:raspberry}}
\caption{Photographs of the embedded system.}
\label{fig:system_photos}
\end{figure}

A Raspberry Pi 3B is used to implement the system (Figure \ref{fig:system_photos}).
%Its small size, computing power and flexibility make it suitable for embedded applications. 
It is not built according to the DO-254/ED-80 standard \cite{hilderman2007avionics} and therefore would not be suitable for installation on an aircraft. However, it is reliable enough to develop an application that will demonstrate the basic functions and performance of this runway overrun avoidance system. A 7-inch touch screen is attached to visualize the results, as well as to insert the necessary inputs for takeoff and landing. A set of speakers is also connected, for the verbal announcements.

\subsection{Software}
The system uses the Raspbian OS (Debian Linux distribution), while the application is developed in the Qt graphical development platform. All the code concerning the calculation of the distances, as well as the related operations, are developed in C++, due to adaptability and performance reasons. The front-end programming is done in Javascript, as provided by the application development tool. The back-end code of the real time application, is parallelized with the interface code for faster, seamless and instant transitions and updates.

\begin{figure}[!t]
\centering
\includegraphics[width=\linewidth]{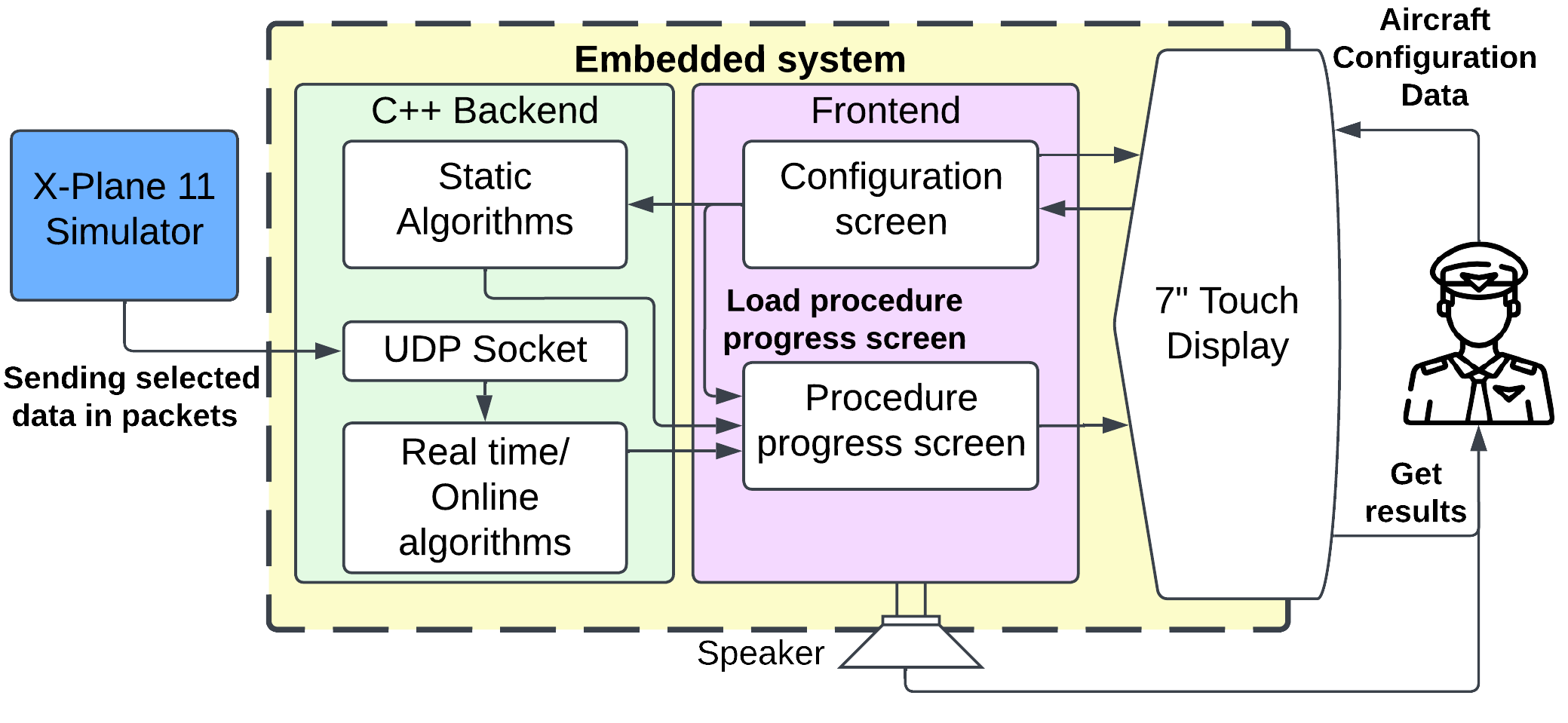}
  \caption{System Architecture.}
  \label{fig:system_architecture}
\end{figure}

The architecture of the system can be seen in Figure \ref{fig:system_architecture}.
Firstly, the pilot inputs all the necessary configuration data through the touchscreen in the configuration screen. These are temporarily saved at the front-end. When clicking on the start button, the new screen (Procedure progress screen) is loaded to the front-end, while all the data are sent to the back-end for the static calculations. As soon as the static calculations are completed, the results are sent to the front-end to display, while the dynamic calculations are triggered. 

During landing or take off, the back-end algorithms continuously unpack the packets that the X-Plane 11 simulator sends to the UDP socket of the RaspberryPi. These are processed by the real time/online algorithms of the back-end and the updated results are sent to the front-end to be displayed. The actual comparison with the length of the remaining runway is done at the front-end.

\begin{figure*}[!t]
\centering

\subfloat[Main menu.]{\includegraphics[width=0.49\columnwidth]{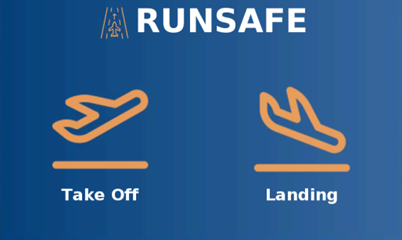}
\label{fig:main_menu}}
\subfloat[Take off configuration screen.]{\includegraphics[width=0.49\columnwidth]{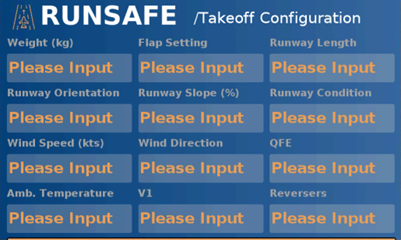}
\label{fig:to_config}}
\subfloat[Flap setting input.]{\includegraphics[width=0.49\columnwidth]{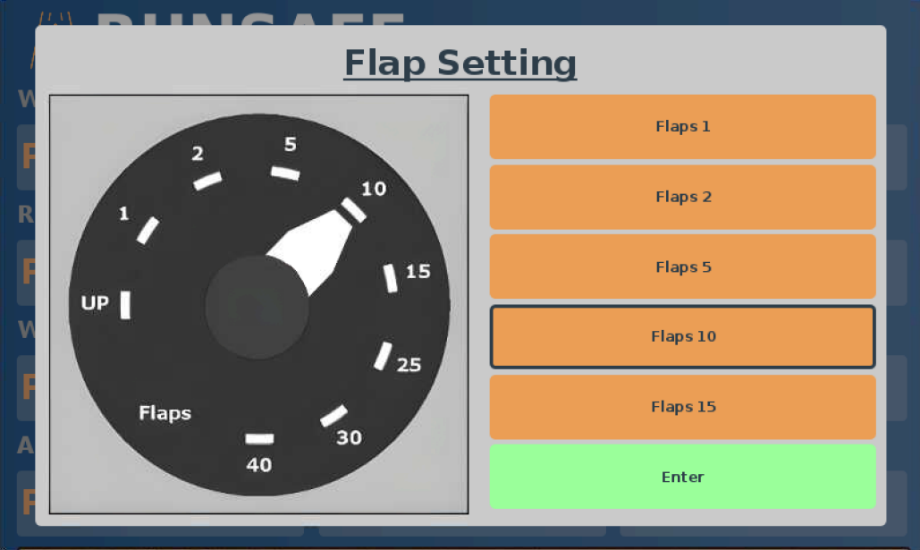}
\label{fig:flap_set}}
\subfloat[Progress screen in warning status.]{\includegraphics[width=0.49\columnwidth]{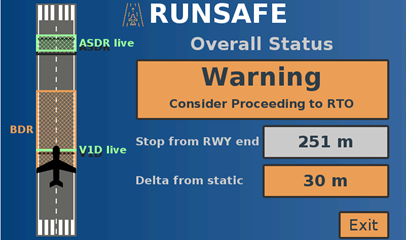}
\label{fig:tO_warning}}
\caption{Screenshots of the application.}
\label{fig:system_in_action}
\end{figure*} 

\begin{figure*}[!t]
\centering
\includegraphics[width=\linewidth]{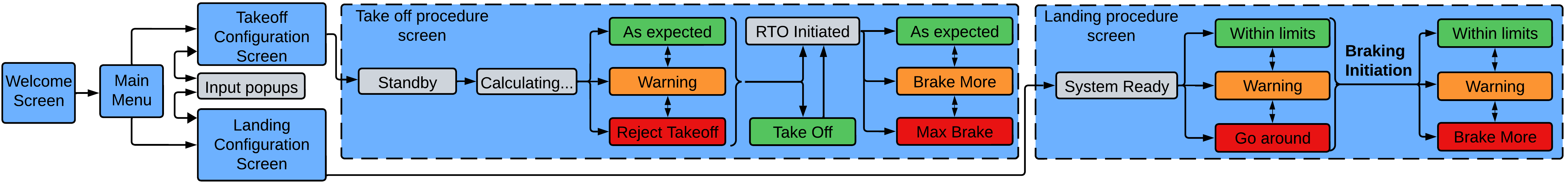}
  \caption{System's screens and states. }
  \label{fig:system_states_and_screens}
\end{figure*}

Figure \ref{fig:system_in_action} shows four indicative screenshots of the application. Figure \ref{fig:main_menu} displays the main menu, that provides the selection between the two procedures, take off and landing and leads to the configuration screen. In Figure \ref{fig:to_config}, the Takeoff Configuration screen is displayed. By selecting one of the fields, a popup window appears and provides either a numpad/keyboard or all the eligible values for this specific input. On the left side, there is a distinct info-graphic display for each input. These, include representations of the aircraft's indications, as for example in the case of the Flap Setting input (Figure \ref{fig:flap_set}), or other means, such as gauges, compasses and others, that provide an actual feedback of the input given.  

Figure \ref{fig:tO_warning}, shows the progress screen of a Take off procedure with a warning prompt. This screen has an identical structure for both procedures. On the left side there is the representation of a runway on which the position of the aircraft is presented in real time, along with the static results (in black color) and the dynamic results (in green color). The take off procedure also provides the distance needed for a Rejected Take Off if it is initiated at this exact speed, indicated as an orange hatched parallelogram named \textit{BDR} (\textit{Braking Distance Required}).   

On the right side of the screen there are useful information about the process's development. The largest of three indicates the overall status of the procedure and changes its background color depending on the current status. \textit{Stop from RWY end} shows how many meters from the runway end the aircraft is expected to come to a complete stop (based on the dynamic calculations, with positive being within the limits), while \textit{Delta from static} indicates the difference between the static and dynamic calculations. The colors of the fields comply with the recommendations of the CS-25 definitions in the cockpit's instruments depending on the severity and the action is needed by the pilots \cite{cs_25_am_27}. Figure \ref{fig:system_states_and_screens} presents the sequence of states and screens of the system.

After the initiation of the Takeoff procedure, the system is in standby mode, waiting for the lifting of the throttle levers. Throughout the spoolup duration, the system is in calculating mode and then it is transitioned to one of the three possible states of dynamic result, named \textit{As expected}, \textit{Warning}, \textit{Reject Take Off}, accompanied with the proper color notation. A warning is raised if the deviation between the static and the dynamic calculations is off by at least 2\%, but the aircraft is not in danger of overrunning the runway. If the latter is true, the system is transitioned to the Reject Take Off state. If the initiation of a rejected takeoff is detected, the system passes to the \textit{RTO} initiated state, before being driven to the dynamic results of the deceleration, \textit{As expected}, \textit{Brake more} and \textit{Max brake}. The criteria of these states, are identical to the previous ones. In the case that the aircraft attains the desired \textit{V1} speed, the system is being driven to the \textit{Take off} status, that indicates that a Rejection should not be initiated, yet it is still capable of detecting one. 

In the case of the Landing procedure, the system is initially at a \textit{System Ready} state, till the aircraft reaches 300ft above the runway level. With dynamic results being available, the system transitions to one of the three available states, \textit{Within limits}, \textit{Warning} or \textit{Go around}, having similar criteria to the Take off procedure. If the aircraft eventually touches the ground and starts braking, a Go Around is no longer an option for safety reasons, thus the three new available states are \textit{Within limits}, \textit{Warning} or \textit{Brake More}. Indicative videos of the system are available online \cite{runsafe_videos}.

\section{Evaluation}
The performance of the system is evaluated through numerous experiments in a wide range of the input parameters. A total of 200 runs was executed for each of the two procedures, designed through a quasi-random, SOBOL sequence Design of Experiments (DOE) that offers uniformly distributed points throughout the configuration space. Results were extracted for both static and dynamic results.

\begin{figure*}[!t]
\centering
\subfloat[Accelerate Stop Distance Required]{\includegraphics[width=0.45\linewidth]{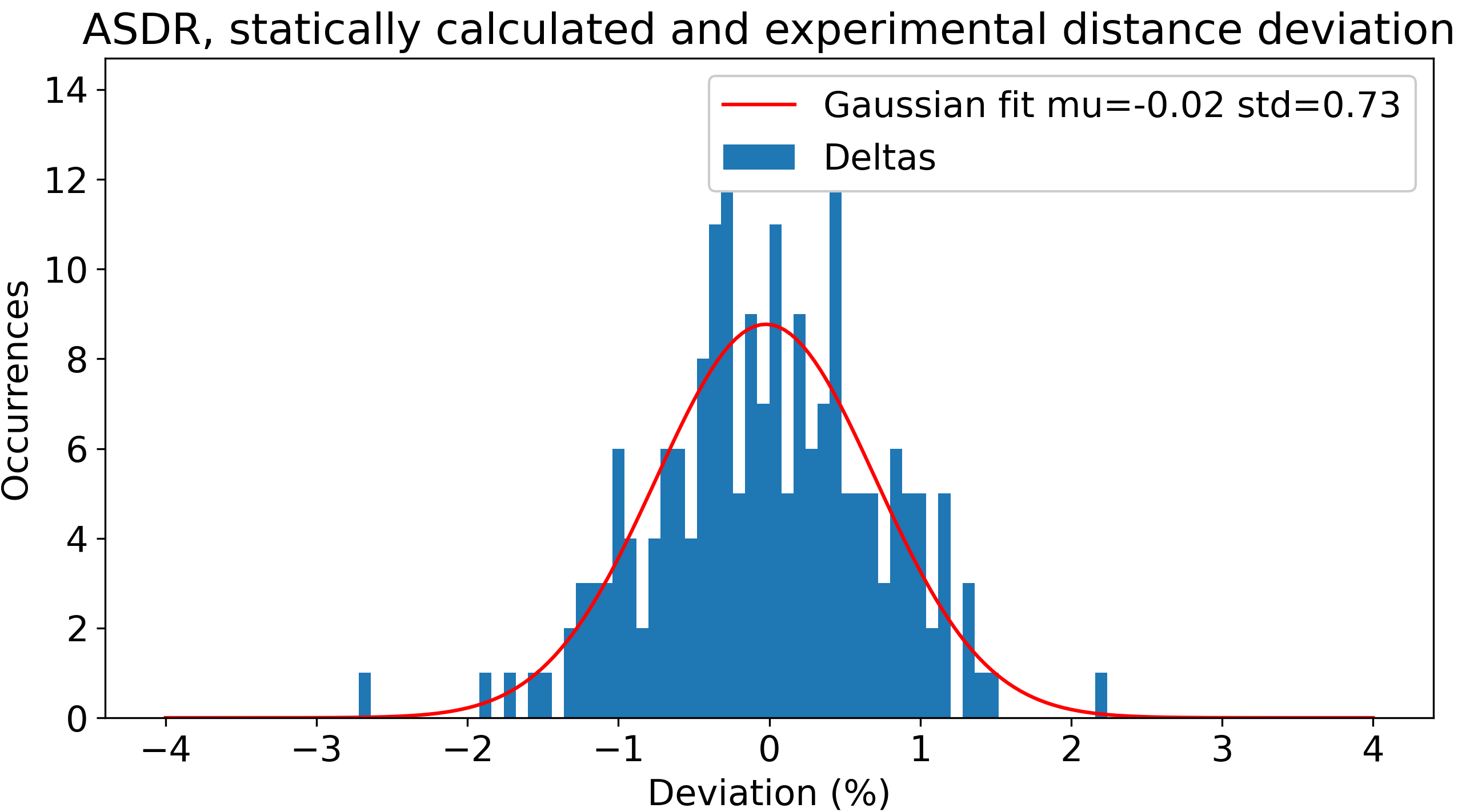}
\label{fig:asdr_static_div}}
\hfil
\subfloat[Landing Distance Required]{\includegraphics[width=0.435\linewidth]{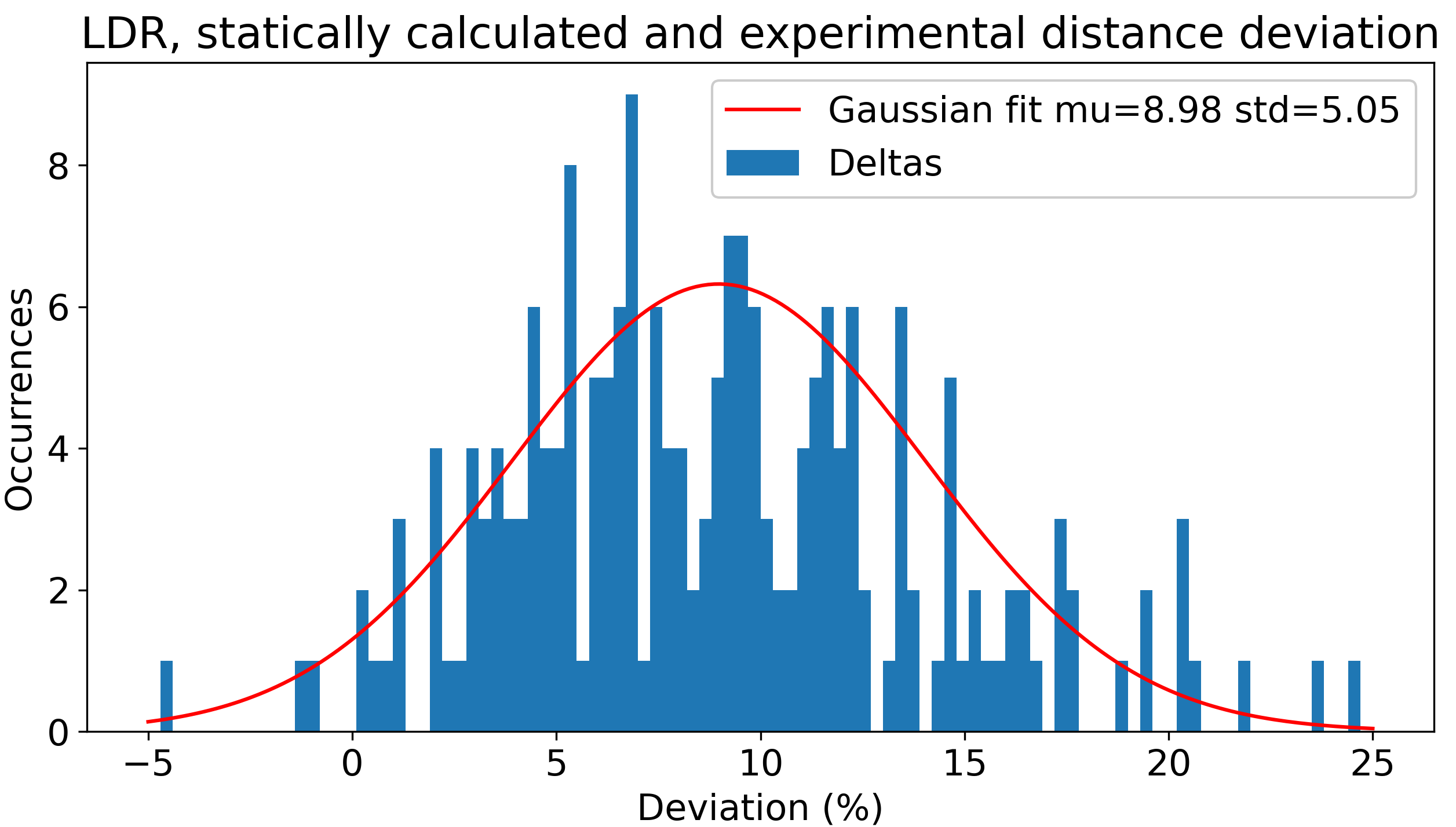}
\label{fig:ldr_static_div}}
\caption{Deviation between the static calculations and experimental results. }
\label{fig:static_divs}
\end{figure*}

\begin{figure*}[!t]
\centering
\subfloat[Acceleration - Take off]{\includegraphics[width=0.32\linewidth]{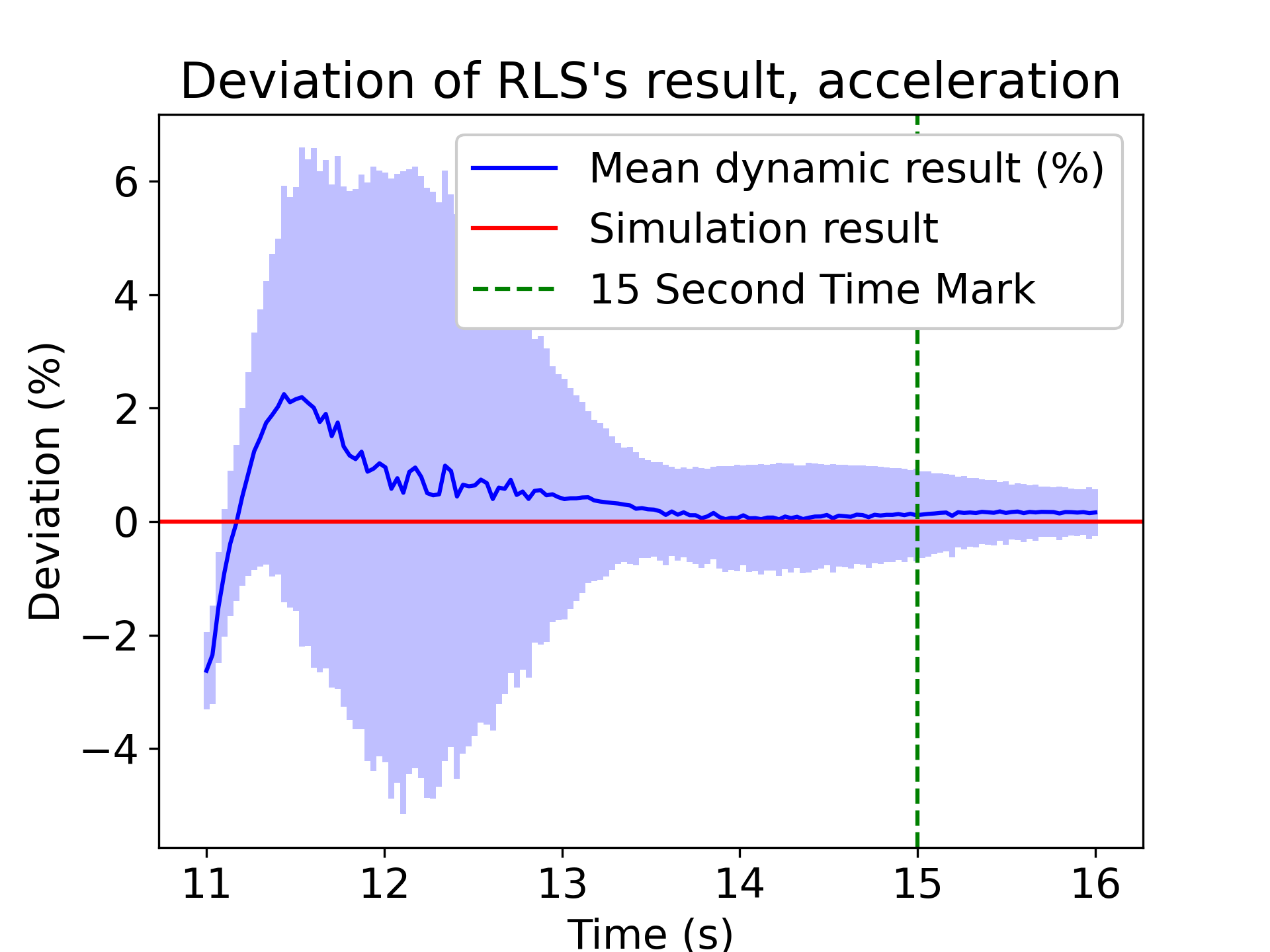}
\label{fig:rls_to_acc}}
\hfil
\subfloat[Deceleration - Take off]{\includegraphics[width=0.32\linewidth]{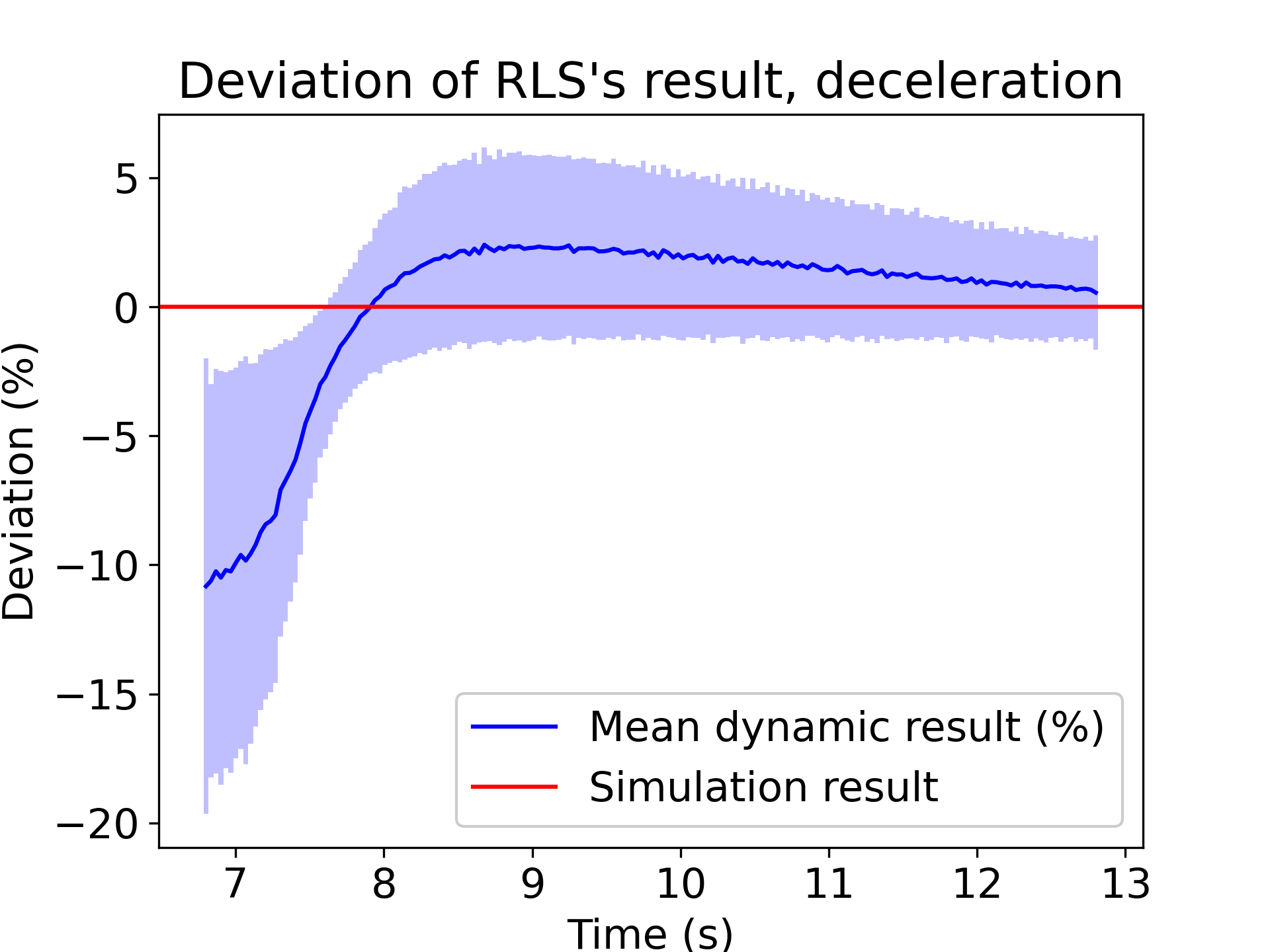}
\label{fig:rls_to_dec}}
\hfil
\subfloat[Deceleration - Landing]{\includegraphics[width=0.32\linewidth]{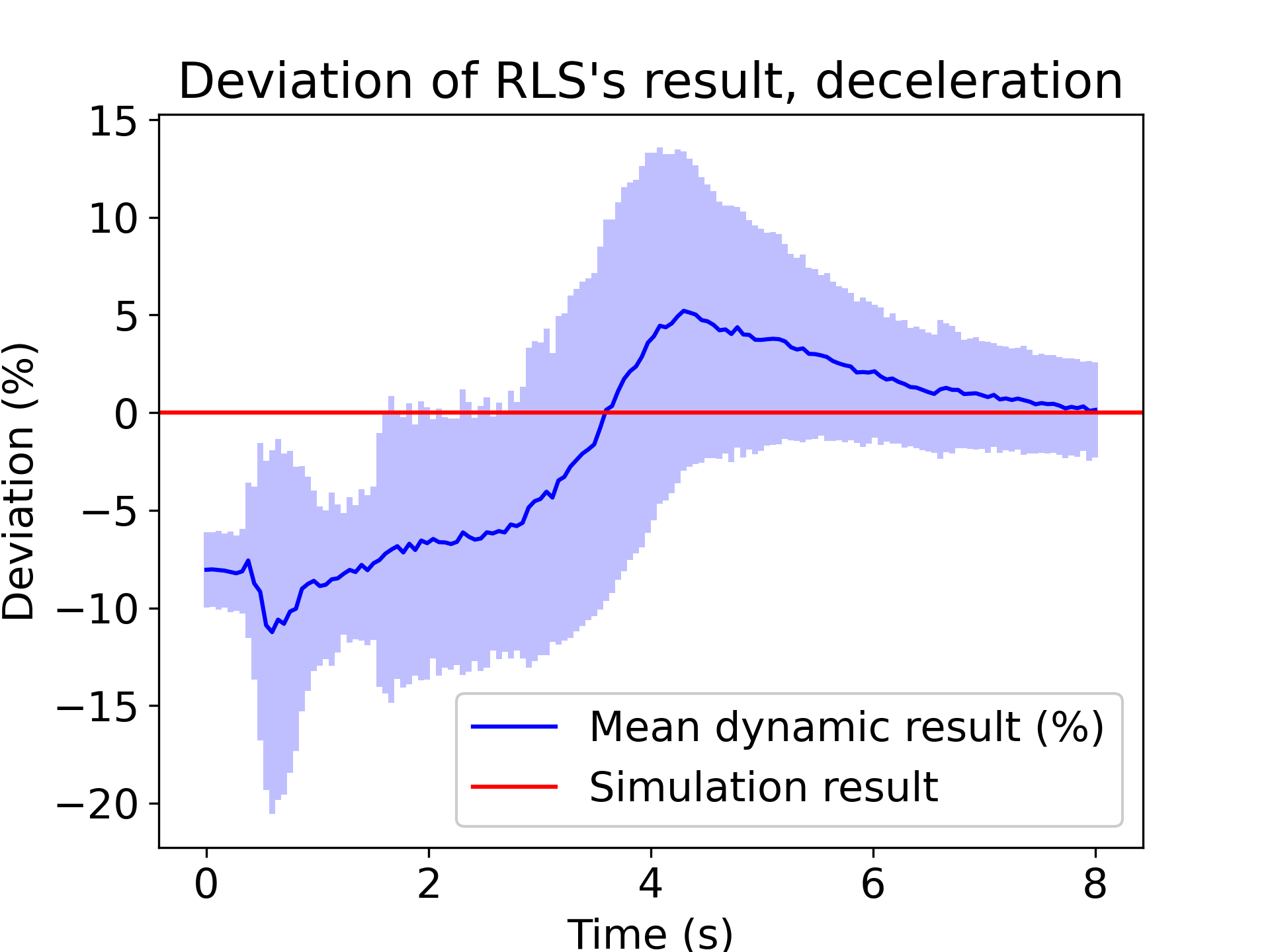}
\label{fig:rls_la_dec}}

\caption{Converge of RLS algorithms in time, based on experimental data.}
\label{fig:rls_dev}
\end{figure*}

\paragraph{Results of the dynamic calculations}

Figure \ref{fig:asdr_static_div} represents the percentile deviations of the statically calculated \textit{Accelerate Stop Distance Required} when compared with the corresponding experimental result. This concerns results of rejected take offs at exactly \textit{V1} speed, without the two second reaction time at speed \textit{V1}, that the CS-25 defines as a safety overestimation parameter \cite{cs_25_am_27}. A histogram of the results and a normal distribution is fitted, with a mean deviation of -0.02\% and a standard deviation of 0.73\%.

On the other hand, the \textit{Landing Distance Required}, is slightly overestimated in each of the individual four stages, as the CS-25 does not specify a certain safety overestimation measure. The histogram and the fitted normal distribution for the \textit{Landing Distance Required} are presented in Figure \ref{fig:ldr_static_div}. The total distance is overestimated by 8.98\% on average, while the standard deviation is 5.05\%.

%%%%%%%%%%%%%%%%%%%%%%%%%
%Dynamic calculations
%%%%%%%%%%%%%%%%%%%%%%%%%

\paragraph{Results of the dynamic calculations}
In Figure \ref{fig:rls_dev}, the mean convergence of the three RLS algorithm variants are displayed. During acceleration, a safe and converged result is provided 15 seconds after the initiation of the take off. Similarly, when braking under a \textit{Rejected Take Off}, the result steadily converges 8.5 seconds after the initiation of the rejection for a reliable result, while during landing after 4 seconds into braking.

%%%%%%%%%%%%%
%New addition - embedded system performance  
%%%%%%%%%%%%%
\paragraph{Embedded system's performance}
After a number of executions, it is revealed than the system needs a mean of 0.139 ms in order to calculate the dynamic acceleration result after receiving a new set of data, with a standard deviation of 0.024ms. The respective times for the dynamic calculations of braking are 0.163 ms and 0.03ms due to the more complex calculations. The whole program uses a mean of 6.9\% of the system's memory and a maximum of 47.5\% of the Quad core CPU. 

\section{Conclusion}
Based on EASA's Research Agenda and accident statistics, it is evident that there is space for improvement in terms of runway safety in aviation. RUNSAFE shows promising results about the feasibility of similar systems and their ability to predict and inform the crew members into taking the right decision, at the right time. Future developments include further compliance with the CS-25 specifications and the hardware and software standards, set by the aviation industry. Additionally, the reproduction of real life excursion events in the simulator, will also provide insightful results regarding the system's performance. 

\bibliography{bibliography}{}
\bibliographystyle{IEEEtran}

\end{document}